\documentclass[conference]{IEEEtran}
\ifCLASSINFOpdf
\else
\fi
\usepackage{url}

\usepackage{bm}

\usepackage{comment}
\usepackage{amssymb}
\usepackage{lipsum}
\usepackage{amsmath}

\usepackage{algorithm}
\usepackage[noend]{algpseudocode}
\usepackage{graphicx,float}
\usepackage{adjustbox}
\usepackage{caption}
\usepackage{subcaption}  
\usepackage{wrapfig} 

\usepackage{amsmath,amsfonts,mathtools}
\usepackage{bbold}

\usepackage{xcolor}

\usepackage{cleveref}

\begin{document}
%
\title{Federated Contrastive Learning of Graph-Level Representations}




%
\author{\IEEEauthorblockN{Xiang Li\IEEEauthorrefmark{1},
Gagan Agrawal\IEEEauthorrefmark{2},
Rajiv Ramnath\IEEEauthorrefmark{1} and
Ruoming Jin\IEEEauthorrefmark{4}}
\IEEEauthorblockA{\IEEEauthorrefmark{1}The Ohio State University, Columbus OH 43210, USA}
\IEEEauthorblockA{\IEEEauthorrefmark{2}University of Georgia, Athens GA 30602, USA}
\IEEEauthorblockA{\IEEEauthorrefmark{3}Kent State University, Kent OH 44242, USA}
}


\maketitle

\begin{abstract}

Graph-level representations (and  clustering/classification based on these  
representations) are required in a variety of applications. Examples 
include identifying malicious network traffic,   prediction of 
protein properties, and many others. 
Often, data has to stay in isolated local systems (i.e., 
cannot be centrally shared for analysis)  due to a variety of considerations 
like  privacy concerns, lack of trust between the parties,  regulations, or simply 
because the data is too large to be shared sufficiently quickly.  This points to the need for federated learning for graph-level representations, a topic that has not been explored much, especially in an unsupervised 
setting.

Addressing this problem, this paper presents a new framework we refer to 
as Federated Contrastive Learning of Graph-level Representations (FCLG).   
As the name suggests, our approach builds on contrastive learning. However, what 
is unique is that we apply contrastive learning at two levels. The first application 
is for local unsupervised learning of graph representations. The second level 
is to address the challenge associated with  data distribution variation  (i.e. the 
``Non-IID issue") when combining local models.   Through extensive experiments on the   
downstream task of graph-level clustering, we demonstrate FCLG outperforms baselines 
(which apply existing federated methods on existing graph-level clustering 
methods) with significant margins.

\end{abstract}


\baselineskip=1.04\baselineskip

%
\IEEEpeerreviewmaketitle

\section{Introduction}
\label{sec: intro}

{\em Attributed graphs}  have lately become backbone of countless systems~\cite{Angles2008_GraphDB_models} as  they capture information 
about individual entities (i.e.,  the {\em node features}) as well as the interactions between them  (i.e. the {\em edges}). 
There has also been a  growing interest  in {\em  graph-level representation learning}, 
where the goal is to  learn embedding  of an entire graph.
This   problem has numerous applications,  such as traffic networks analytics ~\cite{pang2021_cgnn,Xing2020_traffic_GCN},  proteins structure studies ~\cite{yang2018_protein_emb,Glig2021_protein_gcn}, molecular property prediction ~\cite{Wale2006_chemical_class,wu2018moleculenet}, drug discovery ~\cite{Gaudelet2021_drug_discovery,Jiang2021_gnn_drug},  sensor networks~\cite{GarciaGarcia2019_TactileGCNAG},  and others. 

\noindent 
{\bf Motivation for Unsupervised Federated Learning:} 
In many machine learning scenarios, including those involving graphs, 
data cannot be stored and analyzed 
centrally due to reasons like privacy concerns, lack of 
trust (e.g., because of competition) among the data owners,  or regulations. Alternatively, in some cases, data  
is too large to be shared, or at least cannot be shared  sufficiently quickly to provide timely analysis results~\cite{Toyotaro2019_fedfinancial,wu2021_fedgnn,zhang2021_subgraph,ke2021_federated}. 
Such scenarios have led to interest within the graph analytics community 
in the  well-known paradigm of {\em Federated  
Learning}~\cite{McMahan2017_FedAvg}. As a background, federated learning involves 
  training 
centralized models  while the data remains  decentralized -- but by sharing 
and updating 
model parameters~\cite{Xie2021_FederatedGC}. 
As specific examples from graph analytics where this approach is required, 
 drug discovery relies on analysis of  molecular structures (graphs). 
However, industry or pharmaceutical research institutes cannot disclose their private data 
due to the regulatory framework they operate under\cite{Xiong2022_drugdiscovery}.   
As another example, attributed graphs built on communication between different
end-points within a (sub-)network has been used to detect and classify 
malware~\cite{busch2021nf}. While it is important to consolidate models 
across multiple networks or administrative domains, sharing the graphs in a 
timely fashion is impossible.  

This paper focuses on {\em unsupervised}   federated learning. Unsupervised 
learning is an alternative to supervised learning, which in general,  and more specifically in the context of graphs,  
requires labelling that ranges from being expensive and/or time consuming 
to  simply being impossible~\cite{sun2020_infograph}. For example, one typical way of generating labels for chemical structures is performing the time-consuming Density Functional Theory (DFT) calculations~\cite{parr_1983_density}.

As a background, to address the challenges associated with attributed graphs, 
researchers have built  on the 
success of Graph Neural 
Networks (GNNs)~\cite{kipf2017semi_GCN}. 
 GNN based models have achieved advanced performance on {\em  graph-level}  representation learning,  especially in a supervised manner~\cite{xu2018_how_powerful_GNN,Zhang2018_graph_class_gcn,Ying_2018_NIPS_HGRL,pang2021_cgnn,GarciaGarcia2019_TactileGCNAG,Jiang2021_gnn_drug,Xing2020_traffic_GCN,Glig2021_protein_gcn}.  
There has also been an  interest in unsupervised learning approaches 
for graph-level representations~\cite{narayanan_2017_graph2vec,hu_2019_strategies,sun2020_infograph}, 
though for the cases when the data is centrally available. 
This includes the works based on 
{\em Graph Kernels}~\cite{prvzulj_2007_biological,orsini_2015_graph} and limited GNN based works~\cite{sun2020_infograph,icml2020_Multi-view_CL}.

\noindent 
{\bf Contributions of this Paper:}
This paper 
presents  a federated and  unsupervised graph-level learning framework. The 
proposed method is inspired by the recent success of Contrastive Learning (CL) in various unsupervised learning tasks~\cite{chen_2020_simple_CL,sun2020_infograph,mnih_2013_learning}. 
To illustrate the type of challenges we need to address -- it is well known 
that  Federated Learning (FL)  performance is hindered by the ``Non-IID" (Independent and Identical Distribution) issue, i.e. data at different sites follows 
different distributions~\cite{Xie2021_FederatedGC}. Thus, both the 
local models and the global model generated simply by averaging parameters 
of these models may be 
far from the  global optima \cite{li2021_model_CFL}. In fact, the standard FL algorithm, FedAvg~\cite{McMahan2017_FedAvg} might simply not be effective 
when dealing with a set of heterogeneous graphs~\cite{Xie2021_FederatedGC}. Previous work~\cite{Xie2021_FederatedGC,chen_2020_fedbe,zhang2020_FURL} has studied the impact 
of  this Non-IID property especially in supervised learning where label information is explicitly utilized during training. In such scenarios,  skewed label distribution on each client is encoded into the respective local model.

Addressing this problem,  we propose \textbf{F}ederated \textbf{C}ontrastive \textbf{L}earning on \textbf{G}raphs (FCLG), a federated graph embedding method with a novel \textit{two-level contrastive learning mechanism}. Specifically, 
the two levels in our approach are: 1) {\em intra-contrasting} 
during local training  on each client or site, specifically where we 
contrast augmented views of encoded graph-level representations to make each individual graph sample more distinguishable from others; 2) {\em inter-contrasting}   between local and global models, 
ensuring that  the global model captures the common patterns underlying graphs from different clients, to achieve better generalization performance than any local model.   
We also show how contrastive learning can be seen as an improvement over the  
approaches derived from the area of Knowledge Distillation (KD)~\cite{lin2020_ensemble_distillation,kim2021_compare}.

The core contributions of this work can be summarized as:

\begin{itemize}
  \item To our best knowledge, FCLG is the first framework to learn graph-level representations in an unsupervised manner under federated settings.
  \item FCLG introduces a novel two-level contrastive mechanism to alleviate issues caused by the non-IID situations that arise in federated learning. 
  \item Through comprehensive empirical studies using four public benchmark datasets for graph-level representation learning, we show that FCLG consistently outperforms other baselines in the downstream graph-level clustering task.
  
\end{itemize}

The remainder of this paper is organized as follows: Section~\ref{sec: related work} introduces related work in federated learning and unsupervised graph-level representation learning. 
Section~\ref{sec: method} describes the architecture of FCLG and the two-level contrastive mechanism. Section~\ref{sec: experiment}, by extensive experiments,  evaluates the
quality of graph embedding from FCLG for  downstream graph-level clustering tasks and 
demonstrates superior performance over other federated learning baselines. 
Section~\ref{sec: empirical} provides an empirical study on several important factors affecting FCLG performance. 
Section~\ref{sec: conclusion} summarizes our conclusions from this research.

\section{Related work}
\label{sec: related work}

\noindent\textbf{Graph-level representation learning} aims to predict graph-level labels based on node features and graph structural information for an entire graph. 
In recent years, GNN based models have achieved great success in supervised graph-level learning, i.e., on the task of  graph classification \cite{xu2018_how_powerful_GNN,Zhang2018_graph_class_gcn,Ying_2018_NIPS_HGRL,GarciaGarcia2019_TactileGCNAG,Xing2020_traffic_GCN,Glig2021_protein_gcn}. 
Considering unsupervised  graph-level learning, 
early studies  established models mainly based on graph kernels that used 
``hand-crafted"  similarity measures between substructures~\cite{NIPS2009_subtree,JMLR2010_graph_kernels,Yanardag2015_DGK,Yang2018_structural_Conv}. Recent GNN based models~\cite{sun2020_infograph,icml2020_Multi-view_CL} gained state-of-the-art results by applying unsupervised contrastive learning mechanism as described below.

\noindent\textbf{Contrastive learning} based methods learn discriminative representations by reducing the distance between positive sample pairs, while enlarging the distance between negative sample pairs. These approaches have achieved excellent 
performance in visual representation learning~\cite{hjelm_2019_DIM,Tian_multi_coding_2020,chen_2020_simple_CL,Sohn_N-pair_loss_objective_2016,contrastiveLoss_wu2018}. SimCLR~\cite{chen_2020_simple_CL} proposed a simple yet powerful instance-wise contrastive objective to differentiate each data instance from others for image classification.
Recently, the contrastive principle has been applied to establish GNN models for graph-level representation learning~\cite{qiu2020_GCC,icml2020_Multi-view_CL,sun2020_infograph}. 

\noindent\textbf{Federated learning}  (FL) is  a distributed learning paradigm  that can train centralized models on decentralized data.   A detailed discussion of FL is beyond the scope of this paper, but we discuss a few closely related efforts. 
In recent years,  FL has been  studied  in the context of  visual representation learning~\cite{chen_2020_fedbe,li2021_model_CFL,zhang2020_FURL,zhuang2021collaborative,he2021fedcv}. Aligned with our interest in using contrastive learning,  Li {\em et al.}~\cite{li2021_model_CFL} proposed a model-level contrastive learning approach aiming to alleviate non-IID issues and achieved advanced image classification results. 
However, FL studies on  graph-level representations remain limited and have so far focused on supervised learning ~\cite{Xie2021_FederatedGC,he2021_spreadgnn}. 

\section{Methodology}
\label{sec: method}

\subsection{Problem Statement }

Suppose there are $K$ separate clients $C_{1}, C_{2},\cdots, C_{K}$, where client $C_{i}$ has a local graph dataset $S^{i}$ containing a set  of graphs. 
Each graph can be represented as $G = (\mathcal{V}, \mathcal{E}, X)$, where $\mathcal{V}$ is the set of $n$ vertices, 

$\mathcal{E}$ is the edge set, $X \in \mathbb{R}^{n \times F}$ is the feature matrix where each node is associated with a $F$-dimensional feature vector. Let $A$ be the $n \times n$ adjacency matrix capturing all edges in the graph $G$. For simplicity, we can also denote each graph as $G(A, X)$. 

Our objective is to learn a model $ \psi_{\omega} : G \to \mathbb{R}^{d} $, 
where $\omega$ is the corresponding parameter set.  
More specifically,  we have a  federated learning problem where our goal is  to learn a parameter set 
$\omega_{g}$ using the  datasets $S \triangleq \bigcup_{i \in [K]}S^{i} $. 
The individual parameter set at the client $C_i$ is denoted as $\omega_{i}$. In other words, 
the   $d$-dimensional graph-level embedding $ \psi_{\omega}(G)$ must be learnt collaboratively from a set of clients, 
each of which stores a distinct set of  graphs,  while
inter-client raw graph data transfer is disallowed.  


As additional background,  we introduce the first basic FL algorithm FedAvg~\cite{McMahan2017_FedAvg} which is commonly used as the starting point for developing more advanced FL frameworks~\cite{Xie2021_FederatedGC}. 
In FedAvg, the global model on the server will aggregate local model parameters transmitted from clients and distribute the aggregated parameters back to all clients.  During each communication round, each client downloads the model from the server $\omega_{i} \gets \omega_{g}^{(t)}$ and trains the model $\omega_{i}$ with its own data $S^{i}$ for $E_{local}$ epochs. Each client $S^{i}$ will send locally updated parameters $\omega_{i}^{(t)}$ to the server, then the global model $\omega_{g}^{(t+1)}$ will be updated by a 
simple weighted average: 

\begin{equation}
    \label{eq:FedAvg_loss}
    \begin{split}
        \omega_{g}^{(t+1)} = \sum_{i=1}^{K}\frac{|S^{i}|}{|S|}\omega_{i}^{(t)}
    \end{split}
\end{equation} 

where $|S^{i}|$ denotes the size of data on the client $S^{i}$ and $|S |$ is the total size of all data samples distributed over all clients. The server will broadcast newly updated parameters $\omega_{g}^{(t+1)}$ to clients for the next round of training.

\subsection{Basic Ideas}    

To solve the problem we have posed, i.e., federated learning of a graph-level 
representation in an unsupervised fashion, we can build on existing 
frameworks like InfoGraph~\cite{sun2020_infograph} and then apply 
an idea like FedAvg described above. However, as we will demonstrate in the experimental evaluation, such combination produces subpar results.

To provide background on our approach, we use the following formalism. 
Let us say we have a  local model $f_{c}$ from a client and the global model $f_{s}$ on the server.   
On the server, we wish $f_{s}$ to encode all samples into a representation where each sample is more distinguishable from others. In other words, the overall model training will push apart $f_{s}(x_{i})$ and $f_{s}(x_{j})$, where different graph samples $x_{i}$, $x_{j}$ can be from different clients.
However, such a goal may not be achieved by solely aggregating local models since a skewed data distribution can be potentially encoded inside each local model $f_{c}$. 

One way of viewing federated learning is that we are trying to distill knowledge from multiple {\em teacher}  models trained on isolated clients into a single global student model on the server. An earlier federated learning  effort~\cite{lin2020_ensemble_distillation} has applied a knowledge distillation (KD) technique for model fusion in image classification tasks. 
A general objective function of KD (as used by these authors~\cite{lin2020_ensemble_distillation}) is the Kullback-Leibler (KL) divergence loss between the softened probability distributions of the teacher models and the student model.

\begin{align}
    \label{eq: KL-loss-KD}
    & l_{KL} \bigg ( \sigma(f_{c}^{t}(x) ), \sigma(f_{s}(x)) \bigg ) \nonumber \\
    &= \sigma(f_{s}(x)) \cdot \bigg (\log \sigma(f_{s}(x)) - \log \sigma(f_{c}^{t}(x)) \bigg )  
\end{align}

where $\sigma(\cdot)$ denotes the softmax function.

KL divergence loss has achieved considerable success, which can be attributed 
to its ability to control the softness of targets via the temperature-scaling hyperparameter $\tau$. Specifically,  utilizing a larger value makes the softmax vectors smooth over latent classes~\cite{kim2021_compare,furlanello2018_born,tang2020_understanding}. 

Moving further, 
it has been observed that the logit matching is positively correlated to the performance improvement in KD~\cite{kim2021_compare}. These authors considered the mean squared error (MSE) between the logit vectors such that the student model can directly learn the logit of the teacher model.  

\begin{align}
    \label{eq: MSE-loss-KD}
    l_{MSE} \bigg ( f_{c}^{t}(x) , f_{s}(x) \bigg ) 
    &= ||  f_{c}^{t}(x) - f_{s}(x) ||_{2}^{2}  
\end{align} 

For the KD loss functions like KL-divergence in Eq.\ref{eq: KL-loss-KD} and MSE in Eq.\ref{eq: MSE-loss-KD}, we are pushing closer the representations $f_{s}(x_{i})$ and $f_{c}(x_{i})$. This alleviates the local drift by aligning the local logits with that of the server model during training.  

In our work, 
what we 
 discover is that the paradigm of  {\em Contrastive Learning}  can help significantly improve both unsupervised graph-level representation on individual sites as well as in federated settings in a unified fashion.
From a knowledge distillation perspective, the inter-contrasting mechanism can be taken as an advanced KD, more specifically an ensembling distillation technique.

\begin{align}
    \label{eq: inter-contrast-KD}
    l_{inter} 
     &= \log \bigg (1 + \exp( f_{c}^{t}(x) \cdot f_{c}^{t-1}(x) - f_{c}^{t}(x) \cdot f_{s}(x) ) \bigg )  
\end{align}

The idea here is that the local drift will be hindered by using more historical information of local training. We not only try to push closer the in-situ local model $f_{c}^{t}(x) $ and $f_{s}(x)$, but also aim to push apart the current representations $f_{c}^{t}(x) $ away from its previous round $ f_{c}^{t-1}(x) $. 

\subsection{Technical Details} 

Prior graph-level learning approaches, such as InfoGraph,  ~\cite{sun2020_infograph,icml2020_Multi-view_CL} are based on mutual information maximization between multi-scale graph structures like graph-level and sub-graph or node-level representations. 
Specifically, they consider the summarized patch representation centered at each node as a {\em positive example}. This positive example is compared with the global representation of the entire graph, based on mutual information. There are also  negative samples 
that arise from  {\em fake}  graphs' local representations.   

In this paper, we produce positive samples of a given graph by {\em  randomized augmentations} --  a powerful and proven approach to produce similar pairs for deep learning models ~\cite{qiu2020_GCC}  that has not yet been 
fully used for learning graph-level representations. 
Basically, we   create augmented similar and dissimilar graphs and 
then apply an instance-wise contrasting objective~\cite{chen_2020_simple_CL}. 

Moreover, in federated learning setting, this approach can  
directly compare graphs available at each client against each other to sharpen the characteristics of each individual graph sample. 
Thus, one of the novel parts of our approach is that we apply contrastive learning at two levels. The first application ({\em intra-contrasting})
is for local  learning of graph representations -- specifically, 
for contrasting augmented views of graphs within each client as described above. 
The  second level ({\em inter-contrasting}) 
is to  contrast between the global model from the server and the local model on each client, 
as already introduced through Equation~\ref{eq: inter-contrast-KD}.

\noindent
{\bf Intra-contrasting Details:} 
As stated previously, at this level we contrast graph-level representations of different graphs from multiple views. 
Suppose we are given graph-level representations $U$ of a batch of graphs. After 
generating the set of 
augmented views $V$, the contrastive mechanism~\cite{chen_2020_simple_CL}  works 
as follows. We denote $u \in U$ as the representation of a single graph in the view $U$ and $v \in V$ as the augmented counterpart of $u$. Under this contrasting strategy, $v$ forms the only positive sample of $u$ and any other graph representation in these two views, i.e. $U \bigcup V$, is regarded as a negative sample. Thus, we get the contrastive loss: 

\begin{equation}
    \label{eq:CL_pairwise_objective}
    \begin{adjustbox}{max width=0.95\columnwidth}
    $
 L(u,v) = \log \big \{ {\sum_{\substack{z \in U \bigcup V \\ z \neq u} } \exp \big \{ \frac{u \cdot z}{\tau} \big \} }/{ \exp \big \{ \frac{u \cdot v}{\tau} \big \} } \big \}
    $
    \end{adjustbox}
\end{equation}

where $\tau$ is the temperature parameter~\cite{chen_2020_simple_CL}.  Suppose we are training the local model on a batch consisting of $B$ graphs, the overall training loss to be minimized is defined as the average agreement $L$ over all positive pairs as follows:

\begin{equation}\label{eq:CL_loss}
    \begin{split}
        l_{intra} = \frac{1}{2B}\sum_{u \in U}[L(u, v) + L(v, u)]
    \end{split}
\end{equation}   

Intra-contrasting will force the local model to capture robust characteristics behind all the graphs stored on each client. The resulting graph embeddings are expected to make each graph sample more distinguishable from others so as to ease the downstream graph-level clustering task. 

\noindent 
{\bf Inter-contrasting Details:}    
In the method described above, each client can only conduct intra-contrasting for its own data. When there is a skewed or non-identical distribution among 
the clients, the lost opportunity of contrasting with data samples from other clients 
can have a significant impact on the generalization performance of the resulting 
model. This is analogous to how, in federated supervised learning, the local model  
can explicitly encode the local label distribution and thereby drift away from the global optimum during training.
As already stated above,  we draw inspiration from a model-level contrasting mechanism proposed by the federated image classification framework proposed in MOON~\cite{li2021_model_CFL}. This work verified that the skewed local data distribution can cause a \textit{drift} in the local updates leading the local model to learn worse representations than the global model. 
We extended their idea to the graph learning domain in order to control local training \textit{drift} and bridge the gap between graph representations learned by the local model and the global model.  This,  when applied in an unsupervised manner, is the second part of our novel 
two-level contrasting framework.   
Rewriting  Equation~\ref{eq: inter-contrast-KD} in our specific context,  our goal is 
to decrease the distance between $U^{t}$ (representations learned by the current local model) and $U_{s}$ (representations learned by the global model) while increasing the distance between $U^{t}$ and $U_{c}^{t-1}$ (representations learned by the local model with parameters from the previous local epoch). Through this process, we pull $U^{t}$ and $U_{c}^{t-1}$ apart to prevent the local model from drifting along skewed subset of data after iterations of local training. Meanwhile, we push $U^{t}$ and $U_{s}$ closer to keep optimizing the generalization performance of the global model. Specifically we minimize the contrastive loss as:

\begin{equation}
    \label{eq: graph-level CL server and client}
    \begin{adjustbox}{max width=0.9\columnwidth}
    $
    l_{inter} = \log\frac{\exp(sim(U^{t}, U_{s})/\tau^{\prime})+\exp(sim(U^{t}, U_{c}^{t-1})/\tau^{\prime})}{\exp(sim(U^{t}, U_{s})/\tau^{\prime})} 
    $
    \end{adjustbox}
\end{equation}

where $sim(\cdot , \cdot)$ is a cosine similarity function and $\tau^{\prime}$ is the temperature for inter-contrasting.  

Next, we  also construct a model variant by contrasting node-level representations $H$ in place of graph-level representations $U$ by minimizing:

\begin{equation}
    \label{eq: node-level CL server and client}
    \begin{adjustbox}{max width=0.9\columnwidth}
    $
    l^{H}_{inter} = \log\frac{exp(sim(H^{t}, H_{s})/\tau^{\prime})+exp(sim(H^{t}, H_{c}^{t-1})/\tau^{\prime})}{exp(sim(H^{t}, H_{s})/\tau^{\prime})} 
    $
    \end{adjustbox}
\end{equation}

where $H^{t}$, $H_{c}^{t-1}$, and  $H_{s}$ are  the node-level counterpart representations of $U^{t}$, $U_{c}^{t-1}$, and $ U_{s}$, respectively. 

\subsection{FCLG: Overall Framework and Algorithm} 

\begin{figure}
    \centering
    \includegraphics[width=0.9\columnwidth]{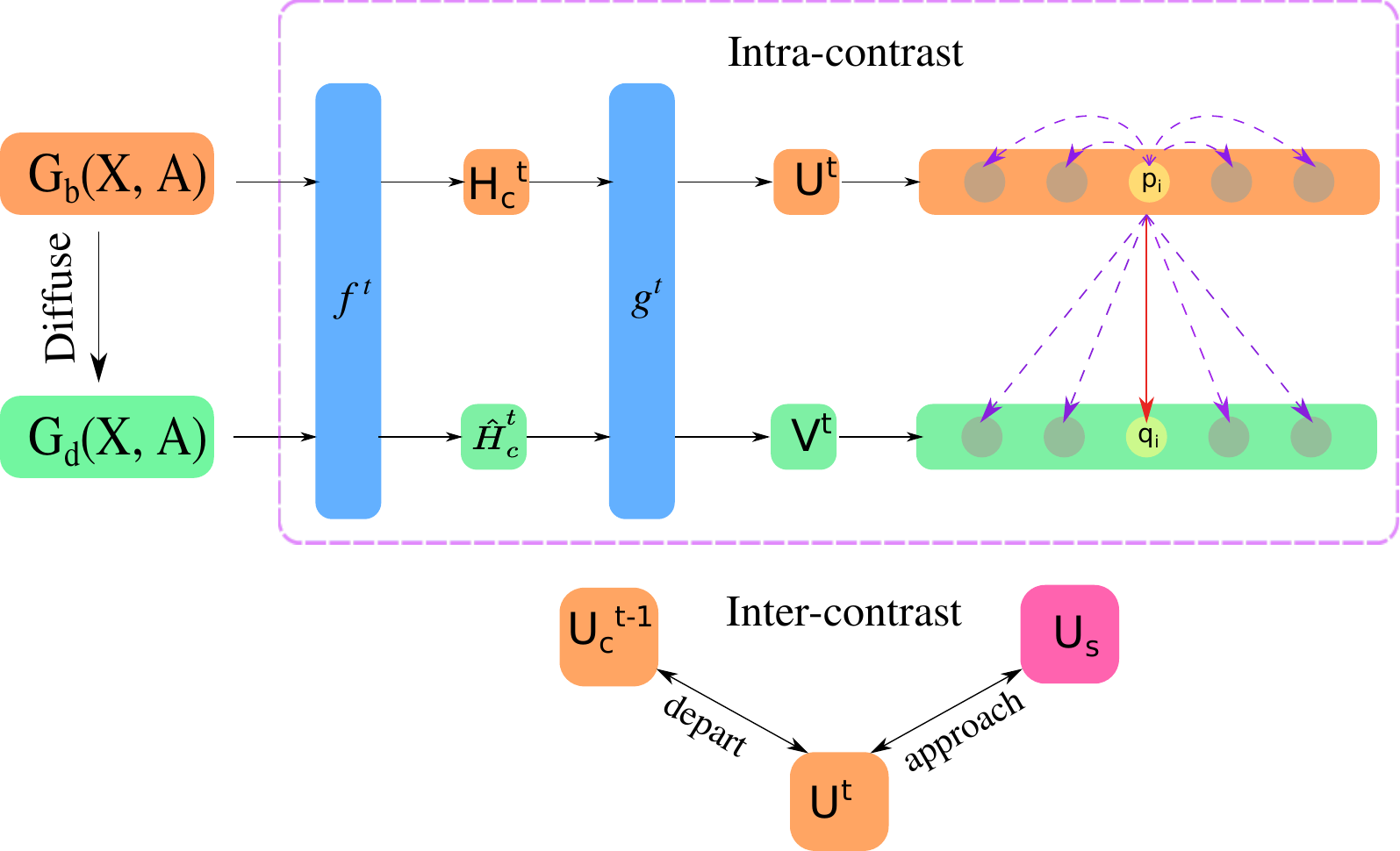}
    \caption{Schematic of FCLG framework architecture. Given one batch of graph data $G_{b}(X, A)$, graph diffusion is used to generate an augmented counterpart $G_{d}(X, A)$.  These two views are fed to the subsequent GIN based encoder $f$ to generate latent node-level representations ($H_{c}^{t}$ and $\widehat{H}_{c}^{t}$,  respectively). The graph pooling function $g$ will further summarize ($H_{c}^{t}$, $ \widehat{H}_{c}^{t}$) by summing up all features for each graph along the columns, yielding $(U^{t},V^{t})$, on which intra-contrastive learning is performed (the superscript $t$ indicates the local training epoch). The inter-contrastive learning step will pull apart the representations ($U_{c}^{t-1}$, $U^{t}$) between adjacent local training epochs on each client, and bring closer  ($U^{t}, U_{s}$) to improve the generalization performance of $U_{s}$. }
    \label{fig:FCLG-arch_two_level}
\end{figure}

\noindent\textbf{Framework Architecture}. Based on the two-level contrastive learning 
idea, our framework first learns a powerful graph embedding in a self-supervised manner from decentralized data. The resulting embedding can then be used for downstream graph-level clustering tasks. 
Figure~\ref{fig:FCLG-arch_two_level} presents an overview of FCLG architecture. Given a batch of input graph data $G_{b}(X, A)$, a graph diffusion precomputing procedure~\cite{klicpera_2019_diffusion}  is performed to generate the  augmentation view $G_{d}(X, A)$ (following~\cite{icml2020_Multi-view_CL}). The subsequent graph embedding generation module comprises a GIN~\cite{xu2018_how_powerful_GNN} based encoder $f$. 
This encoder $f$ involves a set of parameters denoted by  $\omega$ and a non-parametric graph pooling function $g$.  $\omega$ thus also denotes the model parameters of FCLG. 
We use the GIN~\cite{xu2018_how_powerful_GNN} due to its advanced performance in graph-level tasks~\cite{Xie2021_FederatedGC} and apply a summing-up graph pooling function as described in~\cite{icml2020_Multi-view_CL,sun2020_infograph,Ying_2018_NIPS_HGRL,Zhang2018_graph_class_gcn}.
Within this module, $f$ containing stacked $L$-layers of GIN, will first abstract information from graphs into a series of latent representations $ \{h^{k} \}_{k=1}^{L} $. Inspired by~\cite{xu_2018_representation}, we summarize representations at all depths of neural networks through concatenation along  the feature dimension, with the 
goal of  capturing  local information centered at every node:

\begin{equation}
    \begin{adjustbox}{max width=0.95\columnwidth}
    $
        H = CONCAT(\{h^{k} \}_{k=1}^{L}), \quad  U = POOL( H ) \nonumber
    $
    \end{adjustbox}
\end{equation}

Here,  $ H $ indicates the node-level latent representations and $U$ h is the pooled graph-level representation of the input graph $G_b(X, A)$. Similarly, $V$ is the pooled graph-level representation of the augmented view $G_d(X, A)$.
The model will be optimized in a self-supervised manner via contrastive learning objectives.

\noindent\textbf{Complete Algorithm}.
The entire training process is summarized in \textbf{Algorithm \ref{alg:FCLG Train}}. During each communication round, the global model on the server will send model parameters to clients, receive the local model from each client,  and update the global model by Eq.\ref{eq:FedAvg_loss} (on line 2-9). 
During local training, each client $C_i$ first downloads the most updated model parameters from the server (on line 11). Next, we perform the training on each batch of graphs $G_{b}$ locally stored on each client. Augmented view of graphs $G_{d}$ will be generated through graph diffusion mechanism summarized earlier  (on line 14). 
The  GIN~\cite{xu2018_how_powerful_GNN} based encoder $f$ and graph pooling function $g$ abstract graph data into latent graph-level representations (on line 15-16): 
1) $U^{t}$ by the current local model $w_{i}^t$ encoding $G_{b}$; 
2) $V^{t}$ by the current local model $w_{i}^t$ encoding the augmented view $G_{d}$; 
3) $U_{s}$ by the global model $w^{j}$ encoding $G_{b}$; 
4) $U_{c}^{t-1}$ by the previous epoch of local model $w_{i}^{t-1}$ encoding $G_{b}$. 
These obtained representations will be used to compute intra-contrasting and inter-contrasting losses (on line 17).
The total training loss can be computed by summing two contrastive losses up (on line 18) by:

\begin{equation}
    \label{eq:FCLG total loss}
    \begin{adjustbox}{max width=0.9\columnwidth}
    $
    l_{tot} = l_{inter} + l_{intra}
    $
    \end{adjustbox}
\end{equation}

\begin{algorithm}[tbh]
	\caption{FCLG training for graph-level representations} 
	\label{alg:FCLG Train}
	\begin{itemize}
    	\Statex{\textbf{Input}: local training epochs $E$, communication rounds $T$, number of clients $K$ and $\gamma$ the fraction of clients participating in each round.
    	}
    	\Statex{\textbf{Output}: learned model parameters $\omega^{T}$ }
	\end{itemize}
	\begin{algorithmic}[1]
	\State{\textbf{Server:}}
	\State{initialize $\omega^{0}$}
	\For{$j = 0, 1, \cdots, T-1 $} 
            \State{Randomly sample a subset of $K_{\gamma}$ clients}
	    \For{$ i = 1, 2, \cdots, K_{\gamma}$ in parallel}
	        \State{send global model $\omega^{j}$ to $C_{i}$}
	        \State{$\omega_{i}^{j} \gets $ \textbf{Client($i, \omega^{j}$)}}
	    \EndFor
	    \State{$\omega^{j+1} \gets \sum^{K_{\gamma}}_{i}\frac{|S^{i}|}{|S|}\omega^{j}_{i}$}
	\EndFor
	\State{return $\omega^{T}$}
	\State{\textbf{Client($i, \omega^{j}$):}}
    \State{$\omega^{0}_{i} \gets \omega^{j}$}
    \For{$t = 1, 2, \cdots, E $} 
        \For{each batch of graphs $ G_{b} \in S^{i}$}
         \State{$G_{d} \gets $ graph diffusion on $G_{b}$}
            \State{$U^{t} \gets g(f_{\omega^{t}_{i}}(G_{b}))$; $V^{t} \gets g( f_{\omega^{t}_{i}}(G_{d}) )$}
            \State{ $U_{s} \gets g(f_{\omega^{j}}(G_{b})) $; $U_{c}^{t-1} \gets g(f_{\omega_{i}^{t-1}}(G_{b})) $}
            \State{Calculate $l_{intra} $ from Eq.\ref{eq:CL_loss} and $l_{inter}$ from Eq.\ref{eq: graph-level CL server and client}}
            \State{$l_{tot} \gets l_{inter} +  l_{intra} $ }  %
            \State{SGD update on $\omega^{t}_{i}$}
    	\EndFor
	\EndFor
	\State{return $\omega^{E}_{i}$}
	\end{algorithmic} 
\end{algorithm}

\section{Experimental Results}
\label{sec: experiment}

\subsection{Experimental Settings}  
\label{subsec:results0} 

\begin{table}
    \small
    \centering
    \caption{Statistics of datasets for unsupervised graph-level representation learning. 
    }
    \begin{adjustbox}{max width=\columnwidth}
    \begin{tabular}{llrrrr}
    \hline
    Domain & Dataset & Graphs & Avg. Nodes & Avg. Edges & class \\
    \hline
    Proteins & ENZYMES \cite{Karsten2005_protein_graph_kernels}  & $ 600 $ & $ 32.63  $ & $ 62.14 $ & $ 6 $ \\
      & PROTEINS \cite{Karsten2005_protein_graph_kernels}  & $ 1113 $ & $ 39.06 $ & $ 72.82 $ & $  2 $ \\
    Molecules & DHFR \cite{Sutherland2003_spline}  & $ 467 $ & $ 42.43  $ & $ 44.54 $ & $ 2 $ \\
      & NCI1  \cite{Wale2008_chemical} & $ 4110 $ & $ 29.87 $ & $ 32.30 $ & $  2 $ \\
    \hline
    \end{tabular} 
    \end{adjustbox}
    \label{tab: graph representation dataset}
\end{table}

\noindent\textbf{Datasets:}
we use  four datasets~\cite{morris2020_tudataset} for the graph-level clustering task covering multiple domains including: two   protein datasets (ENZYMES, PROTEINS)~\cite{Karsten2005_protein_graph_kernels} and 
two molecule datasets (DHFR~\cite{Sutherland2003_spline} and  NCI1~\cite{Wale2008_chemical}). Each of these datasets comprises   a  set of graphs with binary or multiple class labels. More information about these datasets is presented in \textbf{Table \ref{tab: graph representation dataset}}. We divide graphs of each dataset into 6 clients for federated training.

\noindent 
\textbf{Parameter Settings:} Our FCLG framework is implemented in PyTorch 1.7 on CUDA 10.1, and we use the graph diffusion implementation from Pytorch Geometric~\cite{PyGeom_Fey_Lenssen_2019}. Our experiments are performed on nodes with a dual Intel Xeon 6148s @2.4GHz CPU and dual NVIDIA Volta V100 w/ 16GB memory GPU and 384 GB DDR4 memory. 
We applied a AdamW optimization method with a decoupled weight decay regularization technique \cite{loshchilov2019decoupled_weight_decay}. The detailed hyper-parameter settings are included in the Appendix. 
All baselines are implemented with their officially released codes  as  listed in  
the Appendix 
and we follow their basic settings with all hyper-parameters tuned for optimal clustering performance on each dataset. 
We use the KMeans implementation from scikit-learn \cite{scikit-learn}. 



\noindent\textbf{Baselines:}  Because our work is the first to do unsupervised graph-level 
representation learning in a federated setting we had to construct our own comparison baselines. To this end, 
we combined existing unsupervised graph-level representation methods together with 
popular federated schemes. Specifically, for the former, we used  \textit{InfoGraph}  
that maximizes the mutual information between the graph-level representation and the representations of substructures of different scales~\cite{sun2020_infograph}; \textit{MVGRL}, which involves much of the same idea but contrasts node representations from one view with graph representation of another view~\cite{icml2020_Multi-view_CL}. 
Following similar settings from recent efforts~\cite{Xie2021_FederatedGC,zhang2020_FURL}, we combine different prior federated learning strategies FedAvg and FedProx with the aforementioned graph-level representation learning methods as our baselines. 
Here \textit{FedAvg}~\cite{McMahan2017_FedAvg} aggregates the updated model parameters from local clients and updates the global model by taking average weighted by the dataset size as in Eq.\ref{eq:FedAvg_loss}.
\textit{FedProx}~\cite{li2020federated_prox}, on top of FedAvg, introduces an additional proximal term into training objective to correct the local training updates.

\noindent\textbf{Metrics:} We evaluate the quality of learned graph-level embeddings  
through its performance on the downstream graph-level clustering, specifically 
using two metrics: accuracy and macro F1-score to measure the accordance between prediction and ground-truth labels.
To quantify   non-identical distributions for different (versions of) datasets, 
we apply an average Earthmover's Distance (EMD)~\cite{EMD_rubner1998metric} metric. Following~\cite{hsu_2020_federated}, we take the discrete graph class distribution $ \bf{q_{i}} $ for each client. The class distribution across the total population including datasets from all clients is denoted as $\bf{p}$. Hence, the non-identical distribution  of a dataset is calculated as average distances between  the clients and the population in a weighted fashion:

\begin{equation}
    \begin{adjustbox}{max width=0.95\columnwidth}
    $
        \sum_i\frac{|S_{i}|}{|S|}Dist(\bf{q_{i}}, \bf{p})  \nonumber
    $
    \end{adjustbox}
\end{equation} 

As indicated earlier,  $|S_{i}|$ indicates the data size on each client and $|S|=\sum_{i}|S_{i}|$. Dist$(\cdot,\cdot)$ is a specific distance metric, where in our particular case we use the EMD$(\bf{q}, \bf{p}) \triangleq ||\bf{q} - \bf{p}||_{1}$ and EMD$(\bf{q}, \bf{p}) \in [0, 2]$. Hence a larger EMD implies a more skewed distribution.

\subsection{Graph-level Clustering Results} 
\label{subsec:results1}

\begin{figure}
    \centering
    \includegraphics[width=\columnwidth]{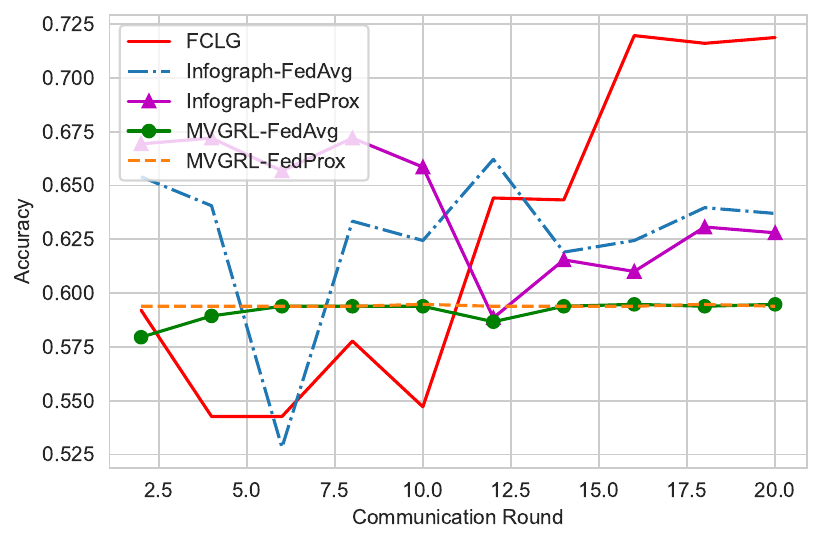}
    \caption{Accuracy Vs. Communication Round on PROTEINS}
    \label{fig:FCLG-acc_commu}
\end{figure}

\begin{table*}[tbh]
    \small
    \centering
    \caption{FCLG graph-level clustering performance for Non-IID client data -- Sections~\ref{subsec:results0} and ~\ref{subsec:results2} explain all the versions compared.  Graph data is  split into 6 clients with specific skewed distribution and respective EMD values provided alongside each dataset name. Metric results (\%) are generated from 10 runs of experiments with converged values after specific communication rounds.
}
    \begin{adjustbox}{max width=\textwidth}
    \begin{tabular}{llrrrrr}
    \hline
    Model & Metric & PROTEINS (0.5774) & ENZYMES (1.2667) & DHFR (0.5694) & NCI1 (0.5995) \\
    \hline
    InfoGraph-FedAvg & Accuracy  & $ 62.40 \pm 1.59 $ & $ 26.00 \pm 1.23 $  & $ 56.59 \pm 0.05 $  & $ 57.82 \pm 0.19 $  \\
    & F1-macro  & $ 61.58 \pm 1.19 $  & $ 25.85 \pm 1.34 $ & $ 55.51 \pm 0.05 $ & $ 57.59 \pm 0.20 $ \\
    \hline
    InfoGraph-FedProx & Accuracy  & $ 60.97 \pm 2.39 $  & $ 27.93 \pm 0.15 $  & $ 57.54 \pm 0.00 $  & $ 59.15 \pm 0.56 $ \\
    & F1-macro  & $ 60.47 \pm 2.19 $  & $ 27.89 \pm 0.18 $ & $ 57.35 \pm 0.00 $ & $ 58.85 \pm 0.85 $  \\
    \hline
    MVGRL-FedAvg & Accuracy  & $ 59.35 \pm 0.10 $  & $ 17.33 \pm 0.00 $  & $ 61.11 \pm 0.00 $  & $ 50.07 \pm 0.00 $  \\
    & F1-macro  & $ 37.25 \pm 0.04 $ & $ 6.12 \pm 0.00 $ & $ 38.25 \pm 0.00 $ & $ 33.41 \pm 0.00 $  \\
    \hline
    MVGRL-FedProx  & Accuracy  & $ 59.30 \pm 0.20 $ & $ 17.50 \pm 0.00 $  & $ 61.11 \pm 0.00 $  & $ 50.07 \pm 0.00 $  \\
    & F1-macro  & $ 37.22 \pm 0.08 $  & $ 6.45 \pm 0.00 $ & $ 38.25 \pm 0.00 $ & $ 33.41 \pm 0.00 $  \\
    \hline
    \hline
    FCLG 
    & Accuracy  &  $ 69.90 \pm 2.59 $ & $ 32.45 \pm 0.45 $  & $ 67.01 \pm 0.83 $  &  $ 61.42 \pm 0.07 $ \\
    & F1-macro  &  $ 69.12 \pm 1.97 $  & $ 30.61 \pm 0.40 $ &  $ 66.59 \pm 0.76 $ &  $ 60.03 \pm 0.07 $  \\
    \hline
    FCLG-H
    & Accuracy  & $ 69.95 \pm 2.57 $  & $ 32.03 \pm 1.22 $  & $ 65.38 \pm 0.84 $  &  $ 61.45 \pm 0.06 $  \\
    & F1-macro  & $ 68.98 \pm 2.08 $  & $ 29.76 \pm 1.53 $ & $ 64.93 \pm 0.90 $ & $ 60.05 \pm 0.07 $  \\
    \hline
    Intra-FedAvg 
    & Accuracy  & $ 70.41 \pm 2.35 $  & $ 30.50 \pm 0.07 $  &  $ 60.22 \pm 1.19 $  & $ 61.55 \pm 0.10 $  \\
    & F1-macro  &  $ 69.29 \pm 1.93 $ &  $ 29.96 \pm 0.07 $ &  $ 59.43 \pm 1.18 $ & $ 60.23 \pm 0.10 $ \\
    \hline
    Intra-KL 
    & Accuracy  &  $ 66.58 \pm 1.14 $ & $ 31.00 \pm 0.00 $  & $ 61.44 \pm 0.25 $  &  $ 61.31 \pm 0.14 $ \\
    & F1-macro  &  $ 66.07 \pm 0.99 $  & $ 29.95 \pm 0.00 $ &  $ 58.59 \pm 0.24 $ &  $ 59.95 \pm 0.17 $  \\
    \hline
    Intra-MSE
    & Accuracy  &  $ 64.80 \pm 2.17 $ & $ 29.83 \pm 0.99 $  & $ 61.48 \pm 0.05 $  &  $ 61.52 \pm 0.07 $ \\
    & F1-macro  &  $ 64.72 \pm 2.09 $  & $ 29.24 \pm 1.19 $ &  $ 59.70 \pm 0.04 $ &  $ 60.25 \pm 0.08 $  \\
    \hline
    \end{tabular} 
    \end{adjustbox}
    \label{tab: Non-IID graph-level clustering}
\end{table*}

\begin{table*}[tbh]
    \small
    \centering
    \caption{FCLG graph-level clustering performance for IID client data -- Sections~\ref{subsec:results0} and ~\ref{subsec:results2} explain all the versions compared. Experiments follow the similar settings of Non-IID condition except that local data distribution of each client is controlled to be the same as or close to the population distribution yielding close-to-zero EMD values.}
    \begin{adjustbox}{max width=\textwidth}
    \begin{tabular}{llrrrrr}
    \hline
    Model & Metric  & PROTEINS(0.0022) & ENZYMES (0.0267) & DHFR (0.0035) & NCI1 (0.0006)  \\
    \hline
    InfoGraph-FedAvg 
    & Accuracy  & $ 64.17 \pm 2.54 $  & $ 26.48 \pm 1.38 $  & $ 57.67 \pm 0.00 $  & $ 59.55 \pm 0.18 $  \\
    & F1-macro  & $ 62.28 \pm 1.83 $ & $ 26.23 \pm 1.30 $ & $ 57.49 \pm 0.00 $ & $ 59.48 \pm 0.20 $  \\
    \hline
    InfoGraph-FedProx
    & Accuracy  & $ 61.51 \pm 4.55 $ & $ 27.35 \pm 0.52 $  & $ 57.54 \pm 0.00 $  & $ 59.40 \pm 0.26 $  \\
    & F1-macro  & $ 60.36 \pm 3.47 $  & $ 27.48 \pm 0.57 $ & $ 57.35 \pm 0.00 $ & $ 59.37 \pm 0.25 $  \\
    \hline
    MVGRL-FedAvg 
    & Accuracy  & $ 59.37 \pm 0.08 $  & $ 17.50 \pm 0.00 $  & $ 61.11 \pm 0.00 $  & $ 50.07 \pm 0.00 $  \\
    & F1-macro  & $ 37.25 \pm 0.03 $  & $ 6.45 \pm 0.00 $ & $ 38.25 \pm 0.00 $ & $ 33.41 \pm 0.00 $  \\
    \hline
    MVGRL-FedProx
    & Accuracy  & $ 59.39 \pm 0.00 $  & $ 17.50 \pm 0.00 $  & $ 61.11 \pm 0.00 $  & $ 50.07 \pm 0.00 $  \\
    & F1-macro  & $ 37.26 \pm 0.00 $ & $ 6.45 \pm 0.00 $ & $ 38.25 \pm 0.00 $ & $ 33.41 \pm 0.00 $  \\
    \hline
    \hline
    FGCL
    & Accuracy  & $ 70.30 \pm 2.10 $  & $ 30.43 \pm 0.13 $  & $ 61.57 \pm 0.78 $  & $ 61.24 \pm 0.37 $  \\
    & F1-macro  & $ 69.07 \pm 1.54 $   & $ 29.89 \pm 0.12 $ & $ 61.00 \pm 0.89 $ & $ 59.66 \pm 0.51 $  \\
    \hline
    FCLG-H 
    & Accuracy  & $ 69.50 \pm 1.70 $ & $ 30.50 \pm 0.00 $  & $ 61.28 \pm 0.84 $  & $ 61.22 \pm 0.00 $  \\
    & F1-macro  & $ 68.65 \pm 1.44 $  & $ 29.98 \pm 0.00 $ & $ 60.59 \pm 1.10 $ & $ 59.71 \pm 0.00 $  \\
    \hline
    Intra-FedAvg
    & Accuracy  & $ 70.69 \pm 0.92 $   & $ 30.33 \pm 0.00 $  &  $ 59.72 \pm 0.84 $  & $ 61.26 \pm 0.02 $  \\
    & F1-macro  &  $ 69.90 \pm 0.71 $ &  $ 29.89 \pm 0.00 $ &  $ 58.68 \pm 0.98 $ & $ 59.76 \pm 0.01 $  \\
    \hline
    Intra-KL
    & Accuracy  &  $ 66.62 \pm 1.31 $ & $ 28.40 \pm 0.85 $  & $ 59.81 \pm 0.10 $  &  $ 61.40 \pm 0.09 $ \\
    & F1-macro  &  $ 66.01 \pm 0.76 $  & $ 28.13 \pm 0.91 $ &  $ 57.04 \pm 0.08 $ &  $ 60.14 \pm 0.09 $  \\
    \hline
    Intra-MSE
    & Accuracy  &  $ 63.87 \pm 2.31 $ & $ 28.47 \pm 1.58 $  & $ 61.77 \pm 0.00 $  &  $ 61.27 \pm 0.13 $ \\
    & F1-macro  &  $ 62.11 \pm 2.56 $  & $ 27.83 \pm 1.51 $ &  $ 59.66 \pm 0.00 $ &  $ 60.11 \pm 0.11 $  \\
    \hline
    \end{tabular} 
    \end{adjustbox}
    \label{tab: IID graph-level clustering}
\end{table*}

\begin{figure*}
     \centering
     \begin{subfigure}[b]{0.33\textwidth}
         \centering
         \includegraphics[width=\textwidth]{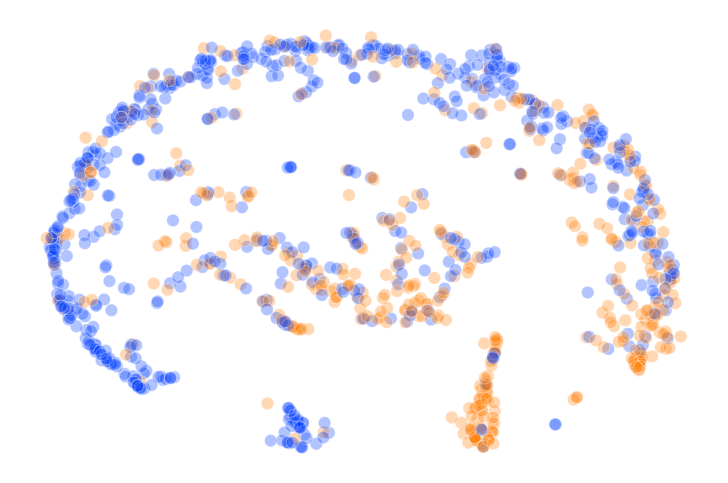}
         \caption{Communication Round 1}
     \end{subfigure}
     \hfill
     \begin{subfigure}[b]{0.32\textwidth}
         \centering
         \includegraphics[width=\textwidth]{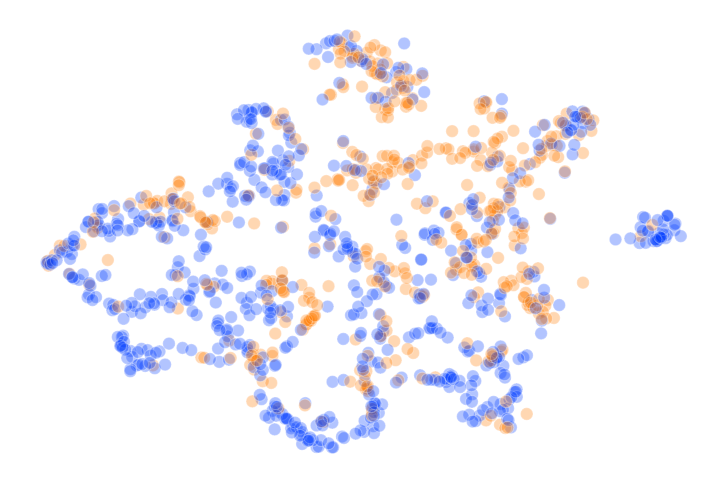}
         \caption{Communication Round 10}
     \end{subfigure}
     \hfill
     \begin{subfigure}[b]{0.32\textwidth}
         \centering
         \includegraphics[width=\textwidth]{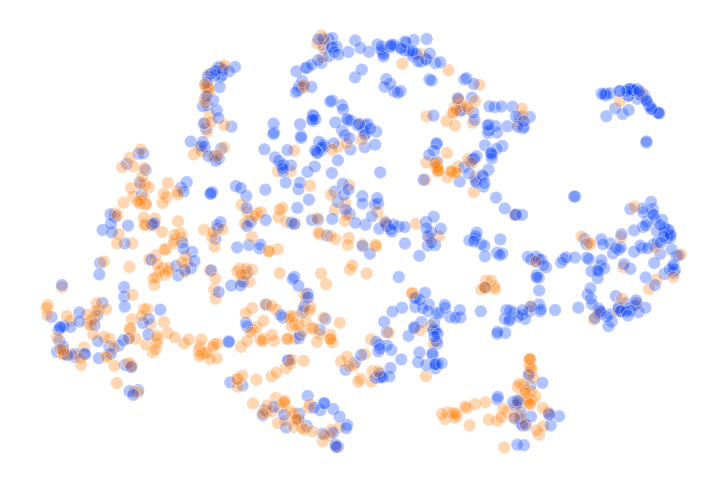}
         \caption{Communication Round 20}
     \end{subfigure}
    \caption{T-SNE for graph-level embeddings of PROTEINS at different communication round during FCLG training. Each color represents a distinct class.}
    \label{fig: FCLG PROTEINS tsne embedding model}
\end{figure*}

We evaluate the clustering performance following the clustering evaluation protocol as taken by previous works \cite{icml2020_Multi-view_CL}.   
Specifically, we set the number of clusters to the number of ground-truth graph-level classes and perform K-Means algorithm \cite{Kmeans_1979} on the resulting graph-level embedding.  
We design two settings based on the data distribution across clients: 1) Non-IID where the graph data distribution over clients will be skewed away from $\bf{p}$, specific EMD values will vary across different data; 2) IID where each client will follow approximately the same population distribution $\bf{p}$ with EMD values close to zero. 

Table~\ref{tab: Non-IID graph-level clustering} shows the graph-level clustering results on four datasets   with non-IID client data -- each experiment is performed for 10 times with both the average 
and the range reported. We provide the EMD values in alignment with each dataset name.
Besides FCLG, we also tested the variant of FCLG (FCLG-H), where intra-contrasting is conducted between node-level representations $H$ via Eq.~\ref{eq: node-level CL server and client}.
Our FCLG based models obtained state-of-the-art graph-level clustering metrics across all datasets.
Specifically, FCLG achieved a significant clustering gain over the closest baseline:  7\% on PROTEINS, around 5\% on ENZYMES and DHFR, and 2\% on NCI1. 
Analyzing further, 
Figure~\ref{fig:FCLG-acc_commu} shows graph-level clustering accuracy after a 
varying number of communication rounds on the dataset PROTEINS. As 
can be seen,  FCLG  starts achieving a clearly superior performance 
after a few rounds. 
The variant we have developed, FCLG-H,  uses node-level representations for inter-contrasting, hence will include more detailed information (and more noise) compared with FCLG. 
The influence of this  appears to be limited since similar clustering performance is observed between FCLG and FCLG-H. Therefore, we can conclude that  FCLG based models show robustness against the scale of graph representations (node-level or graph-level) during inter-contrasting.

In looking across baselines,  we can see MVGRL based models generally have  worse performance. Although MVGRL based models achieve higher accuracy than InfoGraph on DHFR, their macro F1-scores are less  consistent. 
Depending on the neural network applied (i.e. InfoGraph or MVGRL), the different federated strategies (FedAvg and FedProx) have different impacts on clustering performance. 
For MVGRL, both federated strategies yield fairly close clustering results.
For InfoGraph, InfoGraph-FedProx generally performed better (by 1-2\%) than InfoGraph-FedAvg, for ENZYMES, DHFR,  and NCI1.

\subsection{Comparison of FCLG Variants}  
\label{subsec:results2} 

To investigate the effect of different ensembling distillation strategies,  we construct multiple model variants on top of FCLG:  

\begin{itemize} 
\item Intra-FedAvg by removing the term $l_{inter}$ (Eq.\ref{eq:FCLG total loss}) and only keeping intra-contrastive loss; 
\item Intra-KL by replacing $l_{inter}$ with $l_{KL}$ as in Eq.\ref{eq: KL-loss-KD}; 
\item  Intra-MSE by replacing $l_{inter}$ with $l_{MSE}$ as in Eq.\ref{eq: MSE-loss-KD}. 

\end{itemize} 

Our tables~\ref{tab: Non-IID graph-level clustering} ~\ref{tab: IID graph-level clustering} show  results from these versions as well.  First, 
considering Table~\ref{tab: Non-IID graph-level clustering}, 
Several observations can be made. Comparing against the baselines, we can see that our novel intra-contrastive 
learning method is providing significant improvements over the existing centralized baseline approaches. 
Compared with the corresponding FCLG performance, removing  inter-contrasting as in Intra-FedAvg can cause a performance drop in two of the four datasets,  specifically, 
nearly 5\% for DHFR and 2\% for ENZYMES. For PROTEINS and NCI1, no obvious  change  has been observed which indicates that intra-contrasting  with averaging 
itself can yield a robust enough model against local model drifts for certain datasets. 
In comparison with the other models Intra-KL and Intra-MSE, FCLG presents significant advantages, especially in datasets PROTEINS by 3\% and DHFR by 5\%. That is because the KD loss function $l_{inter}$ utilizes richer information (previous round of local model parameters) than either $l_{KL}$ or $l_{MSE}$.

Table~\ref{tab: IID graph-level clustering} covers the graph-level clustering results in one special case where all clients data approximately follow the same distribution (i.e. EMD values close to zero).  
FCLG and FCLG-H continue to outperform the baselines, which in 
this case shows the benefits of our intra-contrasting approach. As in previous 
experiments, the gains are substantial. 
Similar to the observation from the non-IID condition, FCLG has maintained non-trivial advantages over other model variants (Intra-FedAvg, Intra-KL, Intra-MSE) with various distillation techniques.
When compared with non-IID situations, FCLG based models achieve better clustering performance for PROTEINS, are a bit worse metrics for ENZYMES and DHFR, and show similar results for NCI1. 
InfoGraph-FedAvg shows consistently better performance in IID situations than in non-IID ones; however, such consistent trend is not observed in all the other baselines.
Overall, the non-identical distribution effect varies across different combination of models and datasets and a skewed distribution can yield better or worse performance. 
This will be further illustrated by Figure~\ref{fig:FCLG-acc_EMD} as discussed later.

\subsection{T-SNE of Graph-level Classes}

Figure \ref{fig: FCLG PROTEINS tsne embedding model} shows the comparison between graph-level embeddings at the different stages of training using t-SNE algorithm \cite{t-SNE_algo}. We can see that graph-level embedding as obtained from FCLG model can gradually recognize graph-level semantic distributions (i.e.  the overlap reduces between the classes).

\section{Empirical Analysis}
\label{sec: empirical} 


\begin{figure}
     \centering
     \begin{subfigure}[b]{\columnwidth}
         \includegraphics[width=0.95\columnwidth]{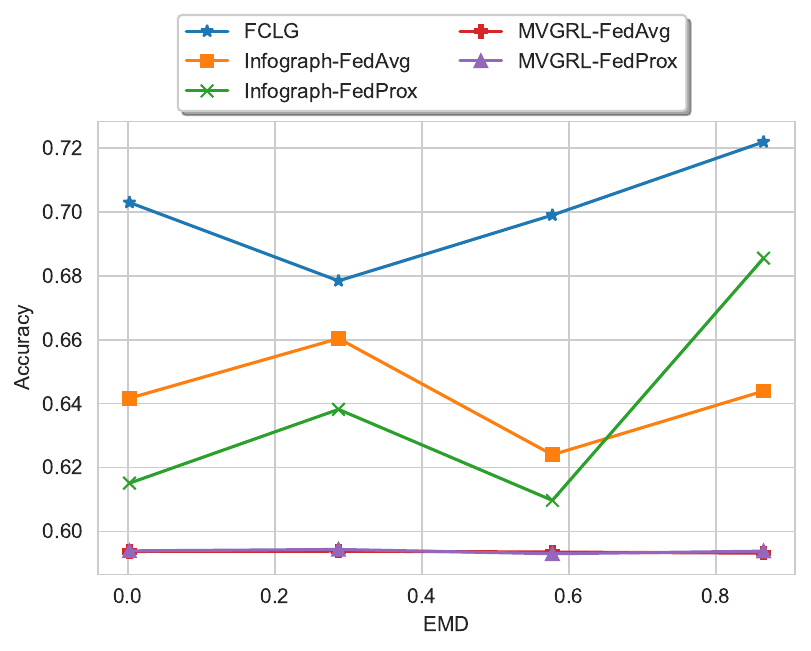}
         \caption{Accuracy Vs. varying skew  of data (fixed number -- 6  clients).}
         \label{fig:FCLG-acc_EMD}
     \end{subfigure}
     \vfill
     \begin{subfigure}[b]{\columnwidth}
         \includegraphics[width=0.95\columnwidth]{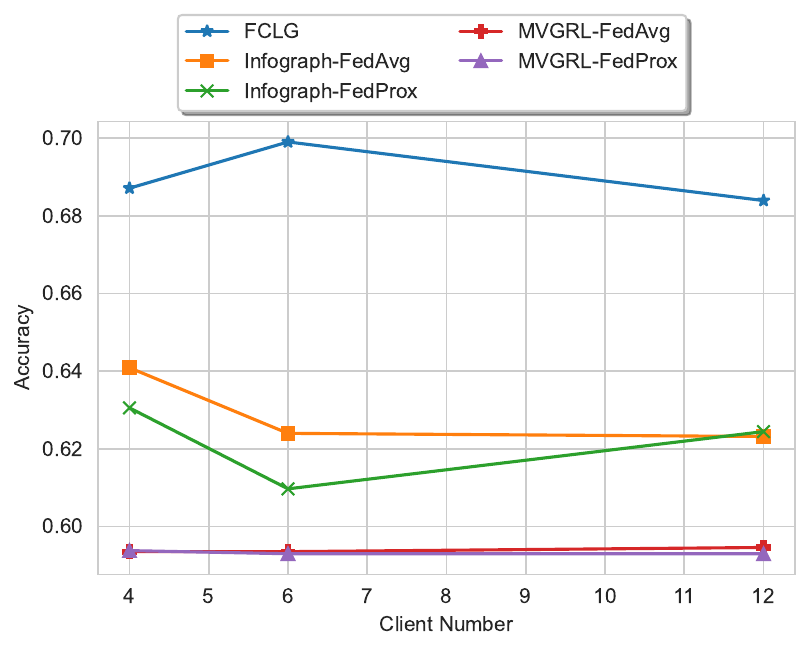}
         \caption{Accuracy Vs. varying number of clients (common EMD value  of 0.58).} 
         \label{fig:FCLG-acc_client_num}
     \end{subfigure}
    \caption{Accuracy Vs. varying data distribution on PROTEINS, each result is obtained by fine-tuning hyper-parameters separately .}
    \label{fig: FCLG accuracy data distribution}
\end{figure}

\begin{figure}
    \centering
    \includegraphics[width=0.95\columnwidth]{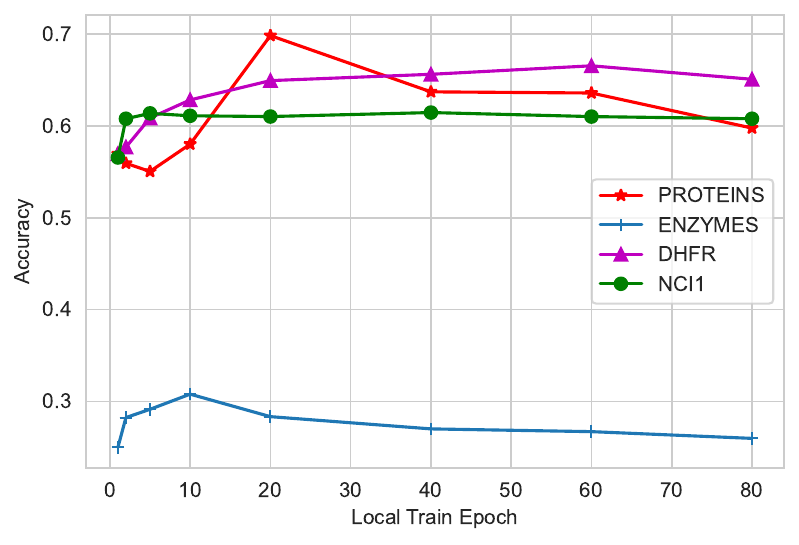}
    \caption{FCLG Accuracy Vs. No. of  Local Train Epoch on PROTEINS (20 communication rounds). Experiments utilize the same Non-IID settings as  in Table~\ref{tab: Non-IID graph-level clustering}. }
    \label{fig:FCLG-acc_local_train_epoch}
\end{figure}

In this section, we study two important 
factors on FCLG's learning power: data distribution across clients, number 
of local training epochs.
Unless explicitly illustrated, experiments for empirical analysis utilize the same Non-IID settings in Table~\ref{tab: Non-IID graph-level clustering} as the default settings and report the converged results averaged from 10 runs.



\subsection{Impact of  Data Distribution}

Figure~\ref{fig: FCLG accuracy data distribution} shows how the  graph-level clustering accuracy changes with different data distributions. 
Specifically, we include two separate experiments:  1) we change data skew 
while the number of  clients  is fixed --   Figure~\ref{fig:FCLG-acc_EMD}; 2) we  vary 
the number of clients while the skew (EMD value) is fixed --  Figure~\ref{fig:FCLG-acc_client_num}.  
Increasing EMD tends to magnify the Non-IID impact while more number of clients can restrict amount of information accessible to each local model.
In both cases, FCLG holds performance superiority over all other baselines  and shows good robustness to both distribution skew  and  the number of clients. 
As expected, 
FCLG's relative performance over baselines is even better with higher EMD value (Figure~\ref{fig:FCLG-acc_EMD}). 
Comparing among the baselines, 
 non-identical distributions do not have a consistent impact on the performance of these baselines and no clear patterns were observed. 
In particular, MVGRL based models  mostly stay constant for different EMD values. 
This observation is different from  those made by earlier   supervised federated learning effort~\cite{Xie2021_FederatedGC,chen_2020_fedbe,zhang2020_FURL}, where increasing 
skew was shown to negatively impact the performance more severely.
This may be because information of labels is not explicitly encoded into models during unsupervised training and thus side effects of skewed graph-level label distribution are lower.

\subsection{Impact of Number of  Local Training Epochs}

We investigate the influence of number of  local training epochs  on graph-level clustering accuracy as in Figure~\ref{fig:FCLG-acc_local_train_epoch}. 
It should be noted that  inter-contrasting will not take effect when the local train epoch number is 1. This is because no previous local training model parameters are available for contrasting. Though different datasets show different behaviors,  we can observe  
the following patterns: 
1) very small number of  local train epochs may yield  slow optimization process resulting in low clustering accuracy; 
2) when the number of  local epochs gets very high on the other hand,  local drift may start to dominate and negatively impact the generalization performance. 
This seems like the case of two of the datasets, i.e.,  PROTEINS and ENZYMES.

\section{Conclusions} 
\label{sec: conclusion}
 \vspace*{-1ex}   
 
In this paper, we have presented and evaluated a new federated learning framework referred to 
as FCLG.  This framework has overcome 
 two difficulties in graph-level representations learning under federated settings: 1) inaccessibility of graph-level ground-truth labels; 2) poor generalization performance of federated learning  when different sites see distinct distributions (the non-IID issue).
With a  unique  two-level contrastive learning mechanism,
FCLG has attained high robustness even with non-IID and obtained high quality graph-level representations.
Our extensive evaluation has shown FCLG achieved state-of-the-art graph-level clustering accuracy compared with the  baselines we constructed, specifically,  2-7\% clustering accuracy gain in non-IID settings and 2-4\% in IID settings.

\bibliographystyle{IEEEtran}
%




\bibliography{references}

\begin{thebibliography}{10}
\providecommand{\url}[1]{#1}
\csname url@samestyle\endcsname
\providecommand{\newblock}{\relax}
\providecommand{\bibinfo}[2]{#2}
\providecommand{\BIBentrySTDinterwordspacing}{\spaceskip=0pt\relax}
\providecommand{\BIBentryALTinterwordstretchfactor}{4}
\providecommand{\BIBentryALTinterwordspacing}{\spaceskip=\fontdimen2\font plus
\BIBentryALTinterwordstretchfactor\fontdimen3\font minus \fontdimen4\font\relax}
\providecommand{\BIBforeignlanguage}[2]{{%
\expandafter\ifx\csname l@#1\endcsname\relax
\typeout{** WARNING: IEEEtran.bst: No hyphenation pattern has been}%
\typeout{** loaded for the language `#1'. Using the pattern for}%
\typeout{** the default language instead.}%
\else
\language=\csname l@#1\endcsname
\fi
#2}}
\providecommand{\BIBdecl}{\relax}
\BIBdecl

\bibitem{Angles2008_GraphDB_models}
\BIBentryALTinterwordspacing
R.~Angles and C.~Gutierrez, ``Survey of graph database models,'' \emph{ACM Comput. Surv.}, vol.~40, no.~1, feb 2008. [Online]. Available: \url{https://doi.org/10.1145/1322432.1322433}
\BIBentrySTDinterwordspacing

\bibitem{pang2021_cgnn}
B.~Pang, Y.~Fu, S.~Ren, Y.~Wang, Q.~Liao, and Y.~Jia, ``Cgnn: Traffic classification with graph neural network,'' 2021.

\bibitem{Xing2020_traffic_GCN}
X.~Ji and Q.~Meng, ``Traffic classification based on graph convolutional network,'' in \emph{2020 IEEE International Conference on Advances in Electrical Engineering and Computer Applications( AEECA)}, 2020, pp. 596--601.

\bibitem{yang2018_protein_emb}
K.~K. Yang, Z.~Wu, C.~N. Bedbrook, and F.~H. Arnold, ``{Learned protein embeddings for machine learning},'' \emph{Bioinformatics}, vol.~34, no.~15, pp. 2642--2648, 03 2018.

\bibitem{Glig2021_protein_gcn}
V.~Gligorijevi{\'{c}}, P.~D. Renfrew, T.~Kosciolek, J.~K. Leman, D.~Berenberg, T.~Vatanen, C.~Chandler, B.~C. Taylor, I.~M. Fisk, H.~Vlamakis, R.~J. Xavier, R.~Knight, K.~Cho, and R.~Bonneau, ``Structure-based protein function prediction using graph convolutional networks,'' \emph{Nature Communications}, vol.~12, no.~1, p. 3168, May 2021.

\bibitem{Wale2006_chemical_class}
N.~Wale and G.~Karypis, ``Comparison of descriptor spaces for chemical compound retrieval and classification,'' in \emph{Sixth International Conference on Data Mining (ICDM'06)}, 2006, pp. 678--689.

\bibitem{wu2018moleculenet}
Z.~Wu, B.~Ramsundar, E.~N. Feinberg, J.~Gomes, C.~Geniesse, A.~S. Pappu, K.~Leswing, and V.~Pande, ``Moleculenet: A benchmark for molecular machine learning,'' pp. 513--530, 2017.

\bibitem{Gaudelet2021_drug_discovery}
T.~Gaudelet, B.~Day, A.~R. Jamasb, J.~Soman, C.~Regep, G.~Liu, J.~B.~R. Hayter, R.~Vickers, C.~Roberts, J.~Tang, D.~Roblin, T.~L. Blundell, M.~M. Bronstein, and J.~P. Taylor-King, ``{Utilizing graph machine learning within drug discovery and development},'' \emph{Briefings in Bioinformatics}, vol.~22, no.~6, 05 2021.

\bibitem{Jiang2021_gnn_drug}
D.~Jiang, Z.~Wu, C.~Hsieh, G.~Chen, B.~Liao, Z.~Wang, C.~Shen, D.~Cao, J.~Wu, and T.~Hou, ``Could graph neural networks learn better molecular representation for drug discovery? a comparison study of descriptor-based and graph-based models.'' \emph{J Cheminform}, vol.~13, no.~1, 02 2021.

\bibitem{GarciaGarcia2019_TactileGCNAG}
A.~Garcia-Garcia, B.~S. Zapata-Impata, S.~Orts, P.~Gil, and J.~G. Rodr{\'i}guez, ``Tactilegcn: A graph convolutional network for predicting grasp stability with tactile sensors,'' \emph{2019 International Joint Conference on Neural Networks (IJCNN)}, pp. 1--8, 2019.

\bibitem{Toyotaro2019_fedfinancial}
\BIBentryALTinterwordspacing
T.~Suzumura, Y.~Zhou, N.~Barcardo, G.~Ye, K.~Houck, R.~Kawahara, A.~Anwar, L.~L. Stavarache, D.~Klyashtorny, H.~Ludwig, and K.~Bhaskaran, ``Towards federated graph learning for collaborative financial crimes detection,'' \emph{CoRR}, vol. abs/1909.12946, 2019. [Online]. Available: \url{http://arxiv.org/abs/1909.12946}
\BIBentrySTDinterwordspacing

\bibitem{wu2021_fedgnn}
C.~Wu, F.~Wu, Y.~Cao, Y.~Huang, and X.~Xie, ``Fedgnn: Federated graph neural network for privacy-preserving recommendation,'' 2021.

\bibitem{zhang2021_subgraph}
K.~Zhang, C.~Yang, X.~Li, L.~Sun, and S.~M. Yiu, ``Subgraph federated learning with missing neighbor generation,'' 2021.

\bibitem{ke2021_federated}
C.~Ke and J.~Honorio, ``Federated myopic community detection with one-shot communication,'' 2021.

\bibitem{McMahan2017_FedAvg}
H.~B. McMahan, E.~Moore, D.~Ramage, S.~Hampson, and B.~A. y~Arcas, ``Communication-efficient learning of deep networks from decentralized data,'' in \emph{AISTATS}, 2017.

\bibitem{Xie2021_FederatedGC}
H.~Xie, J.~Ma, L.~Xiong, and C.~Yang, ``Federated graph classification over non-iid graphs,'' \emph{ArXiv}, vol. abs/2106.13423, 2021.

\bibitem{Xiong2022_drugdiscovery}
Z.~Xiong, Z.~Cheng, X.~Lin, C.~Xu, X.~Liu, D.~Wang, X.~Luo, Y.~Zhang, H.~Jiang, N.~Qiao, and M.~Zheng, ``Facing small and biased data dilemma in drug discovery with enhanced federated learning approaches,'' \emph{Science China. Life sciences}, vol.~65, no.~3, p. 529–539, 2022.

\bibitem{busch2021nf}
J.~Busch, A.~Kocheturov, V.~Tresp, and T.~Seidl, ``Nf-gnn: network flow graph neural networks for malware detection and classification,'' in \emph{33rd International Conference on Scientific and Statistical Database Management}, 2021, pp. 121--132.

\bibitem{sun2020_infograph}
F.-Y. Sun, J.~Hoffman, V.~Verma, and J.~Tang, ``Infograph: Unsupervised and semi-supervised graph-level representation learning via mutual information maximization,'' in \emph{International Conference on Learning Representations}, 2020.

\bibitem{parr_1983_density}
R.~G. Parr, ``Density functional theory,'' \emph{Annual Review of Physical Chemistry}, vol.~34, no.~1, pp. 631--656, 1983.

\bibitem{kipf2017semi_GCN}
K.~N. Thomas and M.~Welling, ``Semi-supervised classification with graph convolutional networks,'' in \emph{ICLR}, 2017.

\bibitem{xu2018_how_powerful_GNN}
\BIBentryALTinterwordspacing
K.~Xu, W.~Hu, J.~Leskovec, and S.~Jegelka, ``How powerful are graph neural networks?'' in \emph{International Conference on Learning Representations}, 2019. [Online]. Available: \url{https://openreview.net/forum?id=ryGs6iA5Km}
\BIBentrySTDinterwordspacing

\bibitem{Zhang2018_graph_class_gcn}
M.~Zhang, Z.~Cui, M.~Neumann, and Y.~Chen, ``An end-to-end deep learning architecture for graph classification,'' ser. AAAI'18/IAAI'18/EAAI'18, 2018.

\bibitem{Ying_2018_NIPS_HGRL}
R.~Ying, J.~You, C.~Morris, X.~Ren, W.~L. Hamilton, and J.~Leskovec, ``Hierarchical graph representation learning with differentiable pooling,'' in \emph{Proceedings of the 32nd International Conference on Neural Information Processing Systems}, ser. NIPS'18.\hskip 1em plus 0.5em minus 0.4em\relax Curran Associates Inc., 2018, p. 4805–4815.

\bibitem{narayanan_2017_graph2vec}
A.~Narayanan, M.~Chandramohan, R.~Venkatesan, L.~Chen, Y.~Liu, and S.~Jaiswal, ``graph2vec: Learning distributed representations of graphs,'' \emph{arXiv preprint arXiv:1707.05005}, 2017.

\bibitem{hu_2019_strategies}
W.~Hu, B.~Liu, J.~Gomes, M.~Zitnik, P.~Liang, V.~Pande, and J.~Leskovec, ``Strategies for pre-training graph neural networks,'' \emph{arXiv preprint arXiv:1905.12265}, 2019.

\bibitem{prvzulj_2007_biological}
N.~Pr{\v{z}}ulj, ``Biological network comparison using graphlet degree distribution,'' \emph{Bioinformatics}, vol.~23, no.~2, pp. e177--e183, 2007.

\bibitem{orsini_2015_graph}
F.~Orsini, P.~Frasconi, and L.~De~Raedt, ``Graph invariant kernels,'' in \emph{Twenty-Fourth International Joint Conference on Artificial Intelligence}, 2015.

\bibitem{icml2020_Multi-view_CL}
K.~Hassani and A.~H. Khasahmadi, ``Contrastive multi-view representation learning on graphs,'' in \emph{Proceedings of International Conference on Machine Learning}, 2020, pp. 3451--3461.

\bibitem{chen_2020_simple_CL}
T.~Chen, S.~Kornblith, M.~Norouzi, and G.~Hinton, ``A simple framework for contrastive learning of visual representations,'' in \emph{International conference on machine learning}.\hskip 1em plus 0.5em minus 0.4em\relax PMLR, 2020, pp. 1597--1607.

\bibitem{mnih_2013_learning}
A.~Mnih and K.~Kavukcuoglu, ``Learning word embeddings efficiently with noise-contrastive estimation,'' \emph{Advances in neural information processing systems}, vol.~26, 2013.

\bibitem{li2021_model_CFL}
Q.~Li, B.~He, and D.~Song, ``Model-contrastive federated learning,'' in \emph{Proceedings of the IEEE/CVF Conference on Computer Vision and Pattern Recognition}, 2021.

\bibitem{chen_2020_fedbe}
H.-Y. Chen and W.-L. Chao, ``Fedbe: Making bayesian model ensemble applicable to federated learning,'' \emph{arXiv preprint arXiv:2009.01974}, 2020.

\bibitem{zhang2020_FURL}
F.~Zhang, K.~Kuang, Z.~You, T.~Shen, J.~Xiao, Y.~Zhang, C.~Wu, Y.~Zhuang, and X.~Li, ``Federated unsupervised representation learning,'' \emph{arXiv preprint arXiv:2010.08982}, 2020.

\bibitem{lin2020_ensemble_distillation}
\BIBentryALTinterwordspacing
T.~Lin, L.~Kong, S.~U. Stich, and M.~Jaggi, ``Ensemble distillation for robust model fusion in federated learning,'' in \emph{Advances in Neural Information Processing Systems}, H.~Larochelle, M.~Ranzato, R.~Hadsell, M.~Balcan, and H.~Lin, Eds., vol.~33.\hskip 1em plus 0.5em minus 0.4em\relax Curran Associates, Inc., 2020, pp. 2351--2363. [Online]. Available: \url{https://proceedings.neurips.cc/paper/2020/file/18df51b97ccd68128e994804f3eccc87-Paper.pdf}
\BIBentrySTDinterwordspacing

\bibitem{kim2021_compare}
T.~Kim, J.~Oh, N.~Kim, S.~Cho, and S.-Y. Yun, ``Comparing kullback-leibler divergence and mean squared error loss in knowledge distillation,'' in \emph{Proceedings of the Twenty-Eighth International Joint Conference on Artificial Intelligence, {IJCAI-21}}.\hskip 1em plus 0.5em minus 0.4em\relax International Joint Conferences on Artificial Intelligence Organization.

\bibitem{NIPS2009_subtree}
N.~Shervashidze and K.~Borgwardt, ``Fast subtree kernels on graphs,'' in \emph{Advances in Neural Information Processing Systems}, Y.~Bengio, D.~Schuurmans, J.~Lafferty, C.~Williams, and A.~Culotta, Eds., vol.~22.\hskip 1em plus 0.5em minus 0.4em\relax Curran Associates, Inc., 2009.

\bibitem{JMLR2010_graph_kernels}
\BIBentryALTinterwordspacing
S.~Vishwanathan, N.~N. Schraudolph, R.~Kondor, and K.~M. Borgwardt, ``Graph kernels,'' \emph{Journal of Machine Learning Research}, vol.~11, no.~40, pp. 1201--1242, 2010. [Online]. Available: \url{http://jmlr.org/papers/v11/vishwanathan10a.html}
\BIBentrySTDinterwordspacing

\bibitem{Yanardag2015_DGK}
P.~Yanardag and S.~Vishwanathan, ``Deep graph kernels,'' in \emph{Proceedings of the 21th ACM SIGKDD International Conference on Knowledge Discovery and Data Mining}, ser. KDD '15.\hskip 1em plus 0.5em minus 0.4em\relax New York, NY, USA: Association for Computing Machinery, 2015, p. 1365–1374.

\bibitem{Yang2018_structural_Conv}
C.~Yang, M.~Liu, V.~W. Zheng, and J.~Han, ``Node, motif and subgraph: Leveraging network functional blocks through structural convolution,'' in \emph{2018 IEEE/ACM International Conference on Advances in Social Networks Analysis and Mining (ASONAM)}, 2018, pp. 47--52.

\bibitem{hjelm_2019_DIM}
D.~Hjelm, A.~Fedorov, S.~Lavoie-Marchildon, K.~Grewal, P.~Bachman, A.~Trischler, and Y.~Bengio, ``Learning deep representations by mutual information estimation and maximization,'' in \emph{ICLR 2019}, 2019.

\bibitem{Tian_multi_coding_2020}
Y.~Tian, D.~Krishnan, and P.~Isola, ``Contrastive multiview coding,'' in \emph{Computer Vision -- ECCV 2020}, 2020.

\bibitem{Sohn_N-pair_loss_objective_2016}
K.~Sohn, ``Improved deep metric learning with multi-class n-pair loss objective,'' in \emph{Advances in Neural Information Processing Systems}, 2016.

\bibitem{contrastiveLoss_wu2018}
Z.~Wu, Y.~Xiong, S.~Yu, and D.~Lin, ``Unsupervised feature learning via non-parametric instance discrimination,'' in \emph{Proceedings of the IEEE Conference on Computer Vision and Pattern Recognition}, 2018.

\bibitem{qiu2020_GCC}
J.~Qiu, Q.~Chen, Y.~Dong, J.~Zhang, H.~Yang, M.~Ding, K.~Wang, and J.~Tang, ``Gcc: Graph contrastive coding for graph neural network pre-training,'' ser. KDD '20, 2020, p. 1150–1160.

\bibitem{zhuang2021collaborative}
W.~Zhuang, X.~Gan, Y.~Wen, S.~Zhang, and S.~Yi, ``Collaborative unsupervised visual representation learning from decentralized data,'' in \emph{Proceedings of the IEEE/CVF International Conference on Computer Vision}, 2021, pp. 4912--4921.

\bibitem{he2021fedcv}
C.~He, A.~D. Shah, Z.~Tang, D.~F.~N. Sivashunmugam, K.~Bhogaraju, M.~Shimpi, L.~Shen, X.~Chu, M.~Soltanolkotabi, and S.~Avestimehr, ``Fedcv: a federated learning framework for diverse computer vision tasks,'' \emph{arXiv preprint arXiv:2111.11066}, 2021.

\bibitem{he2021_spreadgnn}
C.~He, E.~Ceyani, K.~Balasubramanian, M.~Annavaram, and S.~Avestimehr, ``Spreadgnn: Serverless multi-task federated learning for graph neural networks,'' 2021.

\bibitem{furlanello2018_born}
T.~Furlanello, Z.~Lipton, M.~Tschannen, L.~Itti, and A.~Anandkumar, ``Born again neural networks,'' in \emph{International Conference on Machine Learning}.\hskip 1em plus 0.5em minus 0.4em\relax PMLR, 2018, pp. 1607--1616.

\bibitem{tang2020_understanding}
J.~Tang, R.~Shivanna, Z.~Zhao, D.~Lin, A.~Singh, E.~H. Chi, and S.~Jain, ``Understanding and improving knowledge distillation,'' \emph{arXiv preprint arXiv:2002.03532}, 2020.

\bibitem{klicpera_2019_diffusion}
J.~Klicpera, S.~Wei{\ss}enberger, and S.~G{\"u}nnemann, ``Diffusion improves graph learning,'' \emph{arXiv preprint arXiv:1911.05485}, 2019.

\bibitem{xu_2018_representation}
K.~Xu, C.~Li, Y.~Tian, T.~Sonobe, K.-i. Kawarabayashi, and S.~Jegelka, ``Representation learning on graphs with jumping knowledge networks,'' in \emph{International conference on machine learning}.\hskip 1em plus 0.5em minus 0.4em\relax PMLR, 2018, pp. 5453--5462.

\bibitem{Karsten2005_protein_graph_kernels}
K.~M. Borgwardt, C.~S. Ong, S.~Schönauer, S.~V.~N. Vishwanathan, A.~J. Smola, and H.-P. Kriegel, ``Protein function prediction via graph kernels,'' \emph{Bioinformatics}, vol.~21, 06 2005.

\bibitem{Sutherland2003_spline}
J.~J. Sutherland, L.~A. O'Brien, and D.~F. Weaver, ``\BIBforeignlanguage{en}{Spline-fitting with a genetic algorithm: a method for developing classification structure-activity relationships},'' \emph{\BIBforeignlanguage{en}{J. Chem. Inf. Comput. Sci.}}, vol.~43, no.~6, pp. 1906--1915, Nov. 2003.

\bibitem{Wale2008_chemical}
N.~Wale, I.~A. Watson, and G.~Karypis, ``Comparison of descriptor spaces for chemical compound retrieval and classification,'' \emph{Knowledge and Information Systems}, vol.~14, no.~3, pp. 347--375, Mar 2008.

\bibitem{morris2020_tudataset}
C.~Morris, N.~M. Kriege, F.~Bause, K.~Kersting, P.~Mutzel, and M.~Neumann, ``Tudataset: A collection of benchmark datasets for learning with graphs,'' \emph{arXiv preprint arXiv:2007.08663}, 2020.

\bibitem{PyGeom_Fey_Lenssen_2019}
M.~Fey and J.~E. Lenssen, ``Fast graph representation learning with {PyTorch Geometric},'' in \emph{ICLR Workshop on Representation Learning on Graphs and Manifolds}, 2019.

\bibitem{loshchilov2019decoupled_weight_decay}
I.~Loshchilov and F.~Hutter, ``Decoupled weight decay regularization,'' 2019.

\bibitem{scikit-learn}
F.~Pedregosa, G.~Varoquaux, A.~Gramfort, V.~Michel, B.~Thirion, O.~Grisel, M.~Blondel, P.~Prettenhofer, R.~Weiss, V.~Dubourg, J.~Vanderplas, A.~Passos, D.~Cournapeau, M.~Brucher, M.~Perrot, and E.~Duchesnay, ``Scikit-learn: Machine learning in {P}ython,'' \emph{Journal of Machine Learning Research}, vol.~12, pp. 2825--2830, 2011.

\bibitem{li2020federated_prox}
T.~Li, A.~K. Sahu, M.~Zaheer, M.~Sanjabi, A.~Talwalkar, and V.~Smith, ``Federated optimization in heterogeneous networks,'' 2020.

\bibitem{EMD_rubner1998metric}
Y.~Rubner, C.~Tomasi, and L.~J. Guibas, ``A metric for distributions with applications to image databases,'' in \emph{Sixth international conference on computer vision (IEEE Cat. No. 98CH36271)}.\hskip 1em plus 0.5em minus 0.4em\relax IEEE, 1998, pp. 59--66.

\bibitem{hsu_2020_federated}
T.-M.~H. Hsu, H.~Qi, and M.~Brown, ``Federated visual classification with real-world data distribution,'' in \emph{European Conference on Computer Vision}.\hskip 1em plus 0.5em minus 0.4em\relax Springer, 2020, pp. 76--92.

\bibitem{Kmeans_1979}
J.~A. Hartigan and M.~A. Wong, ``A k-means clustering algorithm,'' \emph{Journal of the Royal Statistical Society. Series C (Applied Statistics)}, vol.~28, no.~1, pp. 100--108, 1979.

\bibitem{t-SNE_algo}
L.~Van Der~Maaten, ``Accelerating t-sne using tree-based algorithms,'' \emph{J. Mach. Learn. Res.}, vol.~15, no.~1, p. 3221–3245, Jan. 2014.

\end{thebibliography}





\end{document}